\documentclass[conference]{IEEEtran}
\IEEEoverridecommandlockouts
\usepackage{cite}
\usepackage{amsmath,amssymb,amsfonts}
\usepackage{algorithmic}
\usepackage{graphicx}
\usepackage{textcomp}
\usepackage{xcolor}
\usepackage{hyperref}
\usepackage{cleveref}
\usepackage{enumitem}
\usepackage{colortbl}
\usepackage{xcolor}
\usepackage{booktabs}
\usepackage{adjustbox}
\usepackage[font=footnotesize]{subcaption}
\usepackage[flushleft]{threeparttable}

\hypersetup{
     colorlinks   = true,
     citecolor    = black,
     linkcolor    = black,
     urlcolor = black,
}

\let\orgautoref\autoref

\renewcommand{\autoref}[1]
{%
\def\equationautorefname{Eq.}%
\def\figureautorefname{Fig.}%
\def\subfigureautorefname{Fig.}%
\def\sectionautorefname{Sec.}%
\def\subsectionautorefname{Sec.}%
\def\algorithmautorefname{Alg.}%
\orgautoref{#1}%
}

\begin{document}

\title{Modeling Personalized Difficulty\\of Rehabilitation Exercises Using Causal Trees}

\author{\IEEEauthorblockN{Nathaniel Dennler$^1$, Zhonghao Shi$^1$, Uksang Yoo$^2$, Stefanos Nikolaidis$^1$, Maja Matari\'c$^1$}

\thanks{* This work was supported by a National Science Foundation Graduate Research Fellowship \#DGE-1842487.}
\thanks{$^1$Thomas Lord Department of Computer Science, University of Southern California, Los Angeles, CA, USA. Contact: \texttt{dennler@usc.edu}.} 
\thanks{$^2$Robotics Institute, Carnegie Mellon University, Pittsburgh, PA, USA. }
}

\maketitle

\begin{abstract}
Rehabilitation robots are often used in game-like interactions for rehabilitation to increase a person's motivation to complete rehabilitation exercises. 
By adjusting exercise difficulty for a specific user throughout the exercise interaction, robots can maximize both the user's rehabilitation outcomes and the their motivation throughout the exercise. 
Previous approaches have assumed exercises have generic difficulty values that apply to all users equally, however, we identified that stroke survivors have varied and unique perceptions of exercise difficulty. 
For example, some stroke survivors found reaching vertically more difficult than reaching farther but lower while others found reaching farther more challenging than reaching vertically. 
In this paper, we formulate a causal tree-based method to calculate exercise difficulty based on the user's performance. 
We find that this approach accurately models exercise difficulty and provides a readily interpretable model of why that exercise is difficult for both users and caretakers.  
\end{abstract}

\section{Introduction}
Stroke is a leading cause of serious long-term disability in the United States \cite{tsao2022heart}.
In order to restore mobility, people affected by stroke must engage in long-term neurorehabilitation exercises. Through sustained practice of such exercises, stroke survivors can more easily complete activities of daily living (ADLs), leading to increased quality of life \cite{winstein2019dosage}.
However, motivation is a major barrier to successfully completing rehabilitation exercises. 
Rehabilitation robots can increase motivation to perform neurorehabilitation exercises by physically interacting with users and adapting the difficulty of an exercise throughout the exercise session \cite{gorvsivc2017comparison}. 
Adapting exercise difficulty provides an appropriate challenge for the user without making the exercise too difficult to complete.

\begin{figure}
    \centering
    \includegraphics[width=\linewidth]{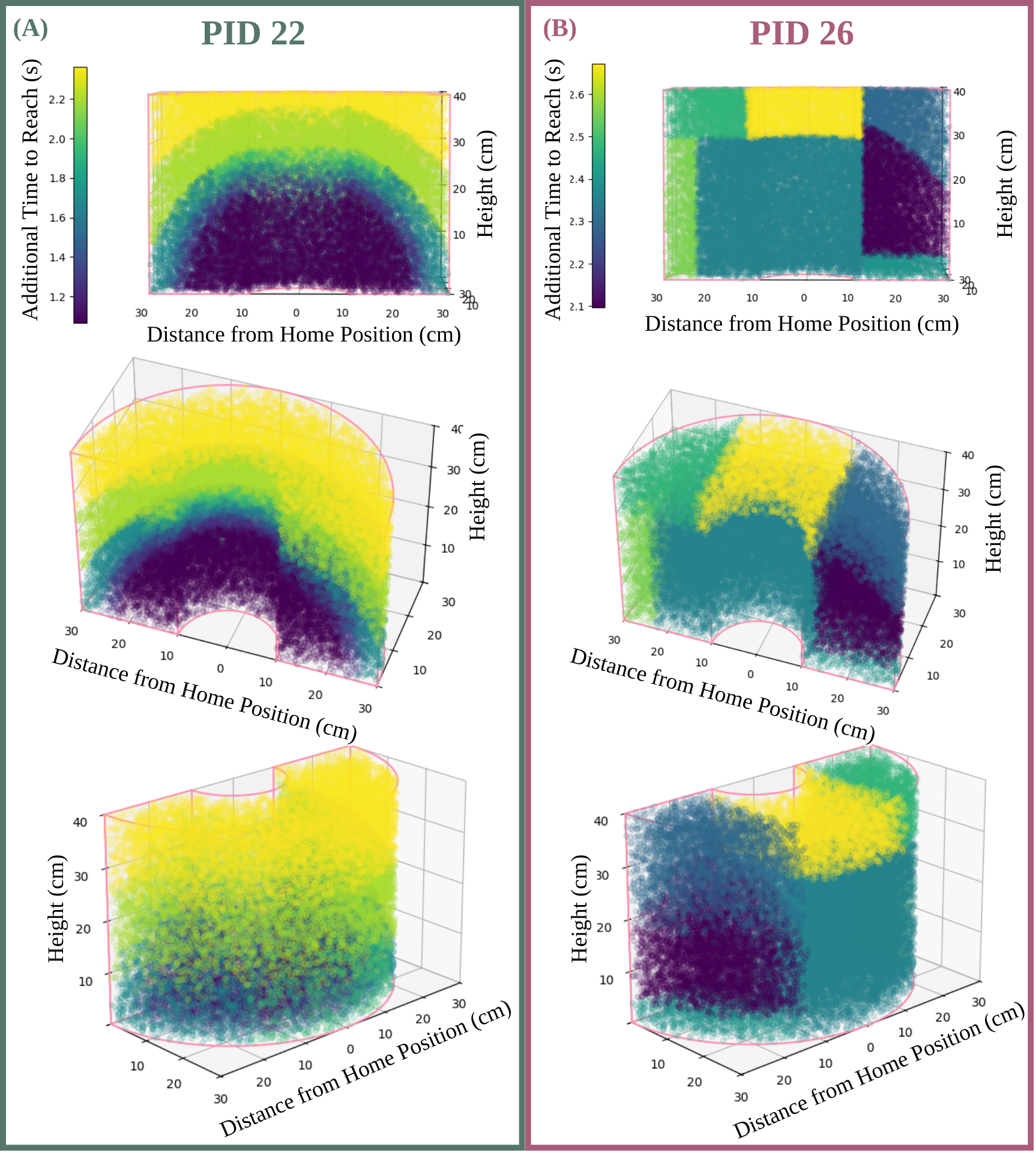}
    \caption{Multiple views of two users' personalized difficulty scores. Lighter colors indicate regions of higher difficulty, determined by statistically significant increases in individual reach times from baseline values.}
    \label{fig:individualized_difficulty}
\end{figure}

Previous algorithms that adapted difficulty throughout the exercise have assumed there is only one controllable variable that changes the difficulty of the exercise, such as the distance to reach \cite{colombo2007design}, speed of a virtual competitor \cite{zimmerli2013increasing}, or the stiffness of a robot resisting motion \cite{metzger2014assessment}. In exercises that reflect daily activities, however, there are many variables that may affect difficulty for a specific user and those variables may vary greatly among people. For example, one stroke survivor may find elbow flexion easier than shoulder flexion, but another may have the opposite experience. To adjust for such differences, clinicians often manually determine exercise difficulty through careful observation of patients \cite{sommerhalder2022armstick}.

Rehabilitation robots have the potential to computationally determine exercise difficulty and subsequently perform these physical exercises. 
Robots can store these exercise completion data and allow both clinicians and users with limited mobility to track rehabilitation progress.
This data collection process provides three key benefits that can aid in rehabilitation.
First, robots can provide objective metrics that are difficult for people to accurately measure, and can communicate those objective metrics to support clinicians and stroke survivors in setting personalized goals \cite{demers2023understanding}.
Second, robots can encourage exercise by providing instruction and social support while reducing social pressure during the exercise \cite{dennler2023metric}.
Finally, robots in homes can provide flexible and consistent access to exercise support for users.

In this work, we present a general algorithm for learning personalized difficulty metrics for rehabilitation exercises from user data. We designed this algorithm to provide explainable reports for both users and clinicians to understand personal difficulty levels, as shown in \autoref{fig:individualized_difficulty}. We evaluate the proposed algorithm using data collected from 15 stroke survivors performing a reaching task with a physically assistive robot arm and a socially assistive robot companion.

\section{Related Work}

Previous work on quantifying exercise difficulty has generally taken one of two approaches. One approach has been to assess the difficulty associated with several specific motor movements in isolation \cite{woodbury2016matching, metzger2014assessment}, while the other approach has used concepts from control theory to assist people throughout entire tasks as needed \cite{asl2017assist,basteris2013adaptive}.

Assessing task difficulty through specific movements in isolation can create a very detailed understanding of personal differences when performing movements such as pinching, elbow flexion, and arm supination \cite{woodbury2016matching}. Past work  used these assessments to initialize difficulty levels for a variety of rehabilitation games \cite{metzger2014assessment}. Assessing individual movements in isolation, however, may not directly transfer to tasks that holistically evaluate combinations of those movements. This is important because activities of daily living involve combinations of many motor functions working together rather than individual motor functions.

In control theory-based approaches, robots adapt in real time to a user to account for the user's level of mobility and other aspects that affect performance, such as fatigue \cite{basteris2013adaptive}. Often, those approaches rely on exo-skeleton-type assistance to obtain a continuous control signal such as torque exerted by the user \cite{asl2017assist}.
In end-effector rehabilitation robots, however, the control signal is typically more sparse. The decision to increase or decrease difficulty can be made after the task is achieved; for example, after an opponent scores in a game \cite{gorvsivc2017comparison} or after the user performs a reaching task \cite{colombo2007design}. In such approaches, parameters of the exercise are increased by a fixed amount, but that approach does not always correspond to a fixed increase in exercise difficulty. 

In this work we aim to augment existing approaches by developing an algorithm that estimates how difficult a task is based on features of that task. Consequently, adaptive rehabilitation exercises can make meaningful difficulty adjustments that correspond to individualized difficulty measurements.

\section{Modeling Personalized Difficulty} \label{sec:causal_tree}

% Many post-stroke rehabilitation programs aim to improve a patient's mobility, seeking to restore function to their pre-stroke capacities, where pre-stroke capacity is often approximated as average neurotypical performance on the exercise. 
% Based on this comparison, we define {\it difficulty} for a stroke survivor to be the difference between their performance and a neurotypical baseline.

The Challenge Point Framework (CPF) \cite{guadagnoli2004challenge} is a motor learning framework that identifies that there is an optimal level of task difficulty to best acheive task improvement. CPF describes two kinds of task difficulties: \textit{nominal task difficulty} and \textit{functional task difficulty}. \textit{Nominal difficulty} refers to the intrinsic difficulty of a task, regardless of the person performing it. \textit{Functional difficulty} refers to difficulty of a task relative to the skill of the person performing the task.

Quantitatively estimating the difference between nominal difficulty and functional difficulty can be challenging due to the high variance in performance measurements for the same task. 
To address this measurement variance, we leverage advances from works that estimate heterogeneous treatment effects from patient covariates \cite{wager2018estimation}. 
Algorithms that estimate treatment effects identify factors that significantly impact outcome measurements, which is highly related to the problem of quantifying task difficulty. At a high level, we learn nominal difficulty by leveraging data collected from neurotypical users completing rehabilitation tasks, and we learn functional difficulty by leveraging the data collected from a specific post-stroke user engaged in the same rehabilitation task.

In order to apply this approach to difficulty estimation, we must make two assumptions. 
First, we assume that exercises have at least one continuous and measurable outcome, for example the time to complete the exercise, that we will denote as $Y$. 
Second, we assume that there are $d$ numerical exercise characteristics that can be modified, e.g., the $(x,y,z)$ location that a user should reach to, that we denote as $X \subset \mathbb{R}^d$. 
We can use the potential outcomes framework \cite{rubin1974estimating} to formalize individual exercise difficulty for task $\vec{x} \in X$ as:

\begin{equation}
    \tau(\vec{x}) = \mathbb{E} \left[ Y^{(1)} - Y^{(0)} | X=\vec{x} \right]
\end{equation}

Where $Y^{(1)}$ denotes the outcome measure of a post-stroke user performing the exercise $\vec{x}$ and $Y^{(0)}$ denotes the outcome measure of a neurotypical user performing exercise $\vec{x}$, i.e., the nominal task difficulty. 
The quantity $\tau(\vec{x})$ denotes the functional difficulty of exercise $\vec{x}$. 

Estimating $\tau(\vec{x})$ directly is challenging because $Y_i$ can be noisy due to unmodeled aspects of the task. 
For example, if the outcome measure is time to completion and a participant is distracted and misses the cue to begin the exercise, they will take longer than usual to reach to the specified $(x,y,z)$ location.
Averaging outcome measures of nearby exercises that the user has performed reduces such noise.

Thus, previous works have found that creating a causal decision tree is a robust technique to determine the nearby points in $X$ that should be included when calculating $\tau(\vec{x})$ \cite{wager2018estimation}. Decision trees iteratively partition $X$ by calculating a splitting rule that separates $X$ into two regions, $X_{left}$ and $X_{right}$. An example splitting rule may be $\vec{x}_z < 0.4 m$, where $X_{left}$ consists of all reaching tasks lower than 0.4 meters and $X_{right}$ consists of all reaching tasks higher than 0.4 meters. The specific rule is selected by finding the exercise characteristic that results in the greatest difference in functional difficulty between $X_{left}$ and $X_{right}$.
Formally, the causal tree learns a function $L(\vec{x})$ that assigns each exercise $\vec{x} \in X$ to a unique set $L$, called a leaf. The estimated functional difficulty for each leaf $L$ is expressed as a difference of averages:

\begin{equation}
    \hat{\tau}(\vec{x}) = \frac{1}{| L|} \sum_{\vec{x} \in L} Y^{(1)} - \frac{1}{|L |} \sum_{\vec{x} \in L} Y^{(0)}
\end{equation}

For more details on the algorithm we used to create the tree, refer to Wagner and Athey \cite{wager2018estimation}. In summary, the causal tree is created using labeled examples from a neurotypical population, $(\vec{x},Y^{(0)})$ and a single post-stroke participant $(\vec{x},Y^{(1)})$. The tree can then be used to predict the functional difficulty $Y^{(1)}$ for a given task $\vec{x}$ for the single post-stroke participant.

\section{User Study}

To evaluate our technique for estimating individualized user difficulty for a reaching exercise. We collected data from 10 neurotypical participants to use as a baseline value, and we collected data from 15 post-stroke participants to asses how well our approach models individual differences in difficulty. The full user study procedure is detailed by Dennler et al. \cite{dennler2023metric} and only briefly summarized here for context. 

\begin{figure}[t]
    \centering
    \includegraphics[width=.8\linewidth]{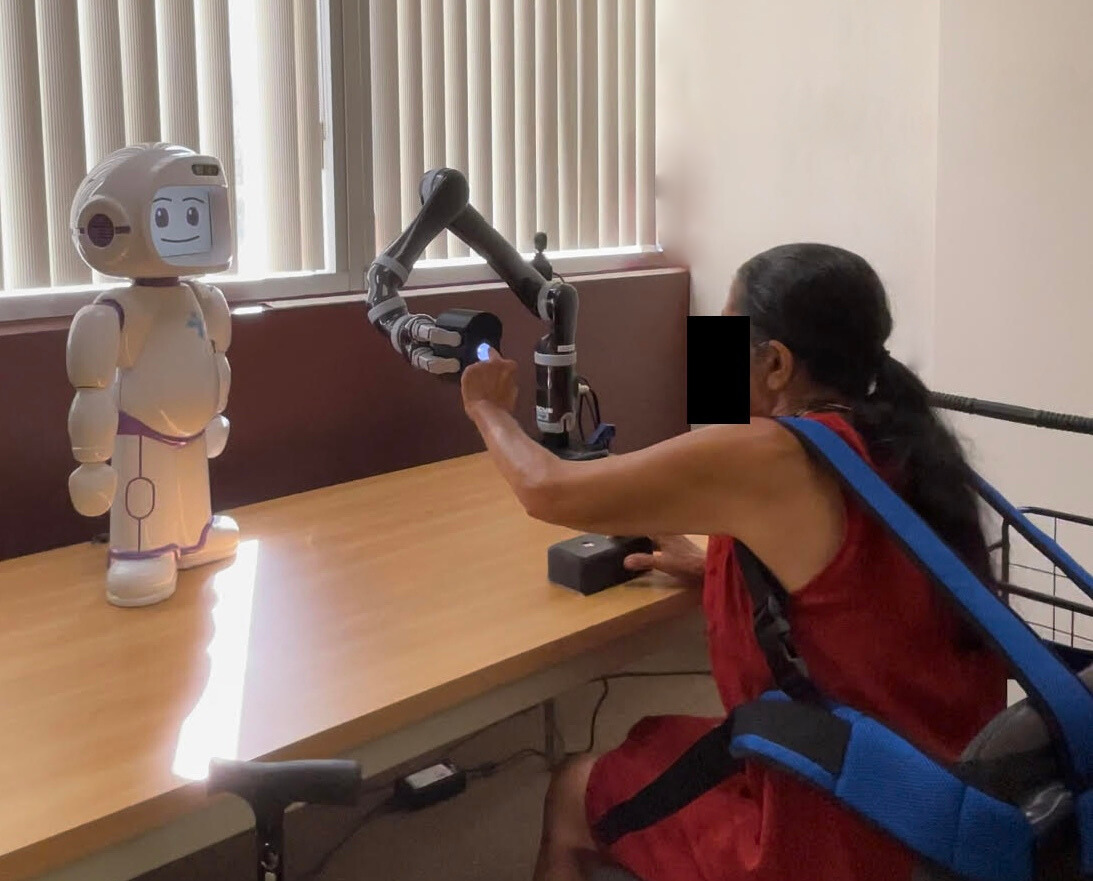}
    \caption{A post-stroke participant performing a reaching exercise. The Socially Assistive Robot (left) provided instructions and feedback. The robot arm (right) moved the reaching target in front of the user. The user reached to the button 100 times using their more affected side.}
    \label{fig:setup}
    \vspace{-1em}
\end{figure}

\subsection{System Description}
Our robotic system that encouraged users to perform reaching exercises, pictured in \autoref{fig:setup}, consisted of a robot arm and a socially assistive robot (SAR). 
The robot arm was the Kinova JACO2 assistive arm selected because it is lightweight and has the similar affordances as end-effector robots typically used for other rehabilitative interactions that have been shown to be effective for neurorehabilitation \cite{lee2020comparisons}. 
The SAR was the Lux AI QTRobot that consisted of a screen face running PyLips \cite{dennler2024pylips}, a 2 degree-of-freedom head, and two 3 degree-of-freedom arms. 
This SAR platform had previously been validated in motivational rehabilitation games with children with cerebral palsy \cite{dennler2021personalizing}. 
The purpose of the SAR was to communicate instructions to the participant, provide positive social reinforcement, and initiate each exercise. This function is similar to previous SAR use in other rehabilitation contexts~\cite{dennler2021personalizing,feingold2021robot}. 

We additionally created two low-cost devices to facilitate the reaching exercise. Both were 3D printed and communicated with the robot system via WiFi using an Adafruit Feather M0 development board. One of the devices was a target object that the Kinova arm held in its end-effector for the users to reach for. The target object consisted of a plastic button with an LED light that turned on to indicate the user should press it. The other device was a home position block for the participant to return to after each reaching exercise. The block consisted of two contact sensors that were 2cm in diameter for participants to rest their left and right pointer fingers. The home position block was placed in front of the midline of the participant and the participant's chair was adjusted so that the participant's elbows formed a right angle when resting their hands on the home position block. The chair also used a harness to minimize compensatory trunk movement during the reaching task.

\begin{figure}[t]
    \centering
    \includegraphics[width=\linewidth]{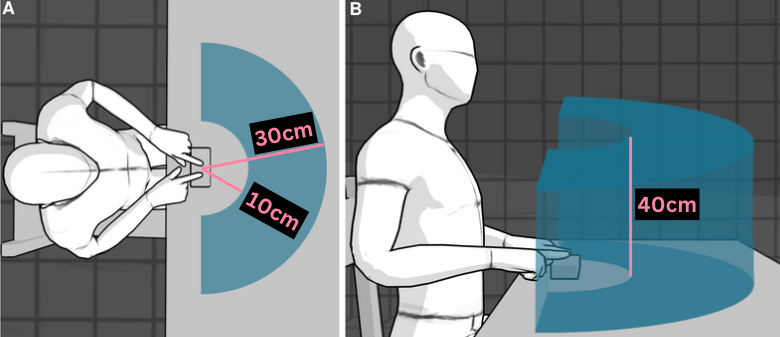}
    \caption{Participants' workspace. The arm moved the target to different points within the blue-shaded region to estimate difficulty.}
    \label{fig:workspace}
    \vspace{-1em}
\end{figure}

\subsection{Procedure}

All procedures were approved by the University of Southern California institutional review board under \#UP-22-00461. Before beginning the exercise session, the experimenter generated a set of 100 evenly-spaced points in a workspace in front of the home position block, visualized in \autoref{fig:workspace}. 
This workspace was defined as a region that extends radially from the home position from 10cm to 30cm in a 180 degree arc. 
The region extended from 0--40cm above the table. The order of these 100 points was then randomized.

Before the participants began the experiment, they signed the consent form. A board-certified physical therapist specializing in neurorehabilitation with more than two years of experience performed the Fugl-Meyer Upper Extremity Motor assessment \cite{fugl1975method} and the experimenter performed the Mini Mental State Exam \cite{folstein1975mini} to determine eligibility. Participants were eligible to participate if they scored higher than 25 out of 66 on the Fugl-Meyer assessment and scored higher than 25 of 30 on the Mini Mental State Exam.

When the participant began the experiment, the SAR provided instructions about the reaching task to inform the participant what would happen. The robot arm then moved to the first randomized point. The light on the target button turned on, and the SAR cued the person to reach to the button by saying ``move", ``ok", ``reach", or ``now". The participant reached to the button with their more affected side. When they pressed the button, the light turned off and the time to reach the button was logged to a file. The SAR periodically provided feedback at random intervals to encourage the participant such as ``Awesome job! That was great.", ``I'm glad to see that you're working hard.", and ``Well done, $<$Participant's Name$>$". The participant returned to the home position and the robot moved to the next randomized point. This was repeated for all 100 points. The SAR updated the participant on their progress at the 25\%, 50\%, and 75\% completion points by saying statements such as ``Three-quarters done! Almost there!".

Each Neurotypical participant completed the procedure once. Each post-stroke participant repeated this procedure up to three times, with at least four days between each consecutive visit. This allowed us to assess the transferability of our approach across sessions.

\subsection{Participants}

We first recruited neurotypical participants to model nominal task difficulty and to adjust any system errors before evaluating on our target population of stroke survivors. This practice helps to minimize the chance of any unforeseen risk to marginalized populations, such as post-stroke participants. All neurotypical participants were right-hand dominant; their demographic information is summarized in \autoref{tab:neurotypical_demographics}. The average age of neurotypical participants was 67 $\pm$ 10 years.

\begin{table}[t!]
\centering
    \caption{Demographic information of the neurotypical group}
    \begin{tabular}{lccc}
    \hline 
         & Median & Minimum & Maximum \\
         \hline
         \vspace{-.9em}\\
         Age (years) & 69.5 & 45 & 82 \\
         Gender & \multicolumn{3}{l}{5 Men, 5 Women}   \\
         Ethnicity &  \multicolumn{3}{l}{2 Asian, 2 Black, 6 White} \\
         \hline
    \end{tabular}
    \label{tab:neurotypical_demographics}
\end{table}

\begin{table}[t]
\centering
\begin{threeparttable}
    \caption{Demographic information of the post-stroke group}
    \begin{tabular}{lccc}
    \hline 
         & Median & Minimum & Maximum \\
         \hline
         % Time from Onset (months) & 100 & 100 & 100 \\
         \vspace{-.9em}\\
         FM-UE Motor Score (66 maximum) & 59 & 27 & 64 \\
         Age (years) & 55 & 32 & 85 \\
         Time between sessions (days) & 6.5 & 4 & 19\\
         Gender & \multicolumn{3}{l}{9 Men, 6 Women}   \\
         Affected Side & \multicolumn{3}{l}{6 Left, 9 Right}   \\
         Ethnicity &  \multicolumn{3}{l}{4 Asian, 3 Black, 4 Hispanic,} \\
         &  \multicolumn{3}{l}{3 White, 1 Mixed-race} \\
         \hline
    \end{tabular}
    \begin{tablenotes}
      \small
      \item Abbreviations: FM-UE, Fugl-Meyer Upper Extremity;
    \end{tablenotes}
    \label{tab:demographics}
    \end{threeparttable}
\end{table}

We found no major issues with the setup after neurotypical participants completed their exercises. We then recruited post-stroke participants to collect data to model functional task difficulty. All post-stroke participants were premorbidly right-hand dominant. Two participants did not meet the study criteria after screening. Thirteen of the remaining fifteen participants completed all three sessions, and two participants completed two sessions due to scheduling constraints. The average age of post-stroke participants was 57 $\pm$ 11 years; demographic information is summarized in \autoref{tab:demographics}. 

\section{Results}

\subsection{Dataset Overview}

We collected two datasets: the neurotypical dataset and the post-stroke dataset. Both datasets used time to reach as the outcome measure, $Y$, and the following numerical exercise characteristics that define each exercise in $X$:
(1) x-position, (2) y-position, (3) z-position, (4) x-position squared---to reflect distance from the participant's midline, (5) distance from home position, and (6) the cue the robot gave to the participant to begin the reaching trial---measured as a categorical variable with four levels.

% \begin{enumerate}
%     \item X position; the distance left or right of the home base, ranging from -0.3m to 0.3m.
%     \item Y position; the distance away from the home base, ranging from 0m to 0.3m.
%     \item Z position; the distance above the table, ranging from 0m to 0.4m.
%     \item Distance; the straight-line distance from the home position to the target point, ranging from 0.1m to 0.5m.
% \end{enumerate}

\subsection{Evaluation Procedure}

\noindent\textbf{Our Method.} We evaluated our method as described in \autoref{sec:causal_tree}. 
First, we concatenated the training datasets of the participant and the neurotypical population. 
We created a group-level variable called ``condition" and marked entries from the participant as ``1" and entries from the neurotypical group as ``0". 
We trained the tree using the econML package \cite{econml2019econml} with the column ``condition" as the treatment variable.

\noindent\textbf{Baselines.} The key advantage of using causal trees is that they reduce variance in each leaf by jointly reasoning about the neurotypical and post-stroke exercise variance, i.e., learning both nominal and functional difficulty \textit{simultaneously}. Thus, we evaluate our approach using baselines that attempt to estimate the outcome for each group \textit{separately} and subtract the two estimates, i.e., the baselines were of the form $f_{s}(\vec{x}) - f_{n}(\vec{x})$, where $f_s$ is a machine learning model that directly estimates the time to reach for the post-stroke participant ($Y^{(1)}$) and $f_n$ is a machine learning model that estimates the time to reach for all neurotypical participants ($Y^{(0)}$).
We evaluated with the following machine learning models as baselines: Decision Tree, Random Forest, Neural Network, Support Vector Machine, k-Nearest Neighbors.

\noindent\textbf{Ground Truth.} We are interested in modeling the functional difficulty of the participant, as this difficulty indicates what exercises should be assigned under the Challenge Point Framework. We define the ground truth functional difficulty for a given participant for the exercise $\vec{x}$ in the workspace as the average time of a single post-stroke participant reaches within a ball with a radius of 5cm centered at $\vec{x}$ minus the average time of all neurotypical participant reaches within the same 5cm ball. Because this value includes data from all of the participant reaches, this serves as an accurate measure of the actual time difference between the participant and the neurotypical group.

\begin{table}[t!]
\caption{Hyperparameters used for our experiments.}
\label{table:hyperparams}
\adjustbox{max width=\linewidth}{%
\begin{tabular}{ rc } 
\hline
Model & Hyperparameters \\
\hline
Causal Forest (Ours)  & \textit{n\_estimators=100, min\_samples=5 }\\
\arrayrulecolor{gray!50}\midrule
\arrayrulecolor{black}
XGBoost & n\_estimators = 200, learning\_rate=0.2, max\_depth=7 \\
Random Forest & n\_estimators=100, max\_depth=100 \\
Neural Network & hidden\_sizes=(50, 50), learning\_rate=1e-3, max\_iter=500 \\ 
SVM & C=10, gamma=0.0001, kernel='rbf'\\
k-Nearest Neighbors & n\_neighbors=15, weights ='distance', metric='manhattan' \\
Decision Tree & max\_depth=10, min\_samples\_split=10, min\_samples=4 \\
\hline
\end{tabular}
}
\end{table}

\begin{figure*}[t]
    \centering
     \begin{subfigure} {.3\linewidth}
      \includegraphics[width=\linewidth]{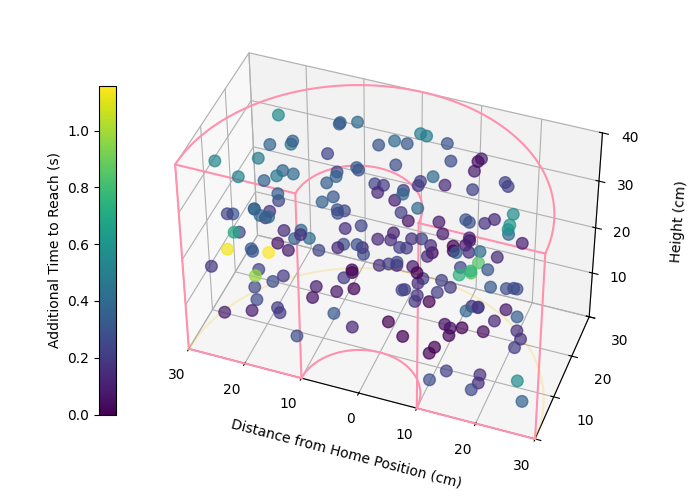}
        \caption{PID 33 ground truth.}
    \end{subfigure}
    \begin{subfigure} {.3\linewidth}
      \includegraphics[width=\linewidth]{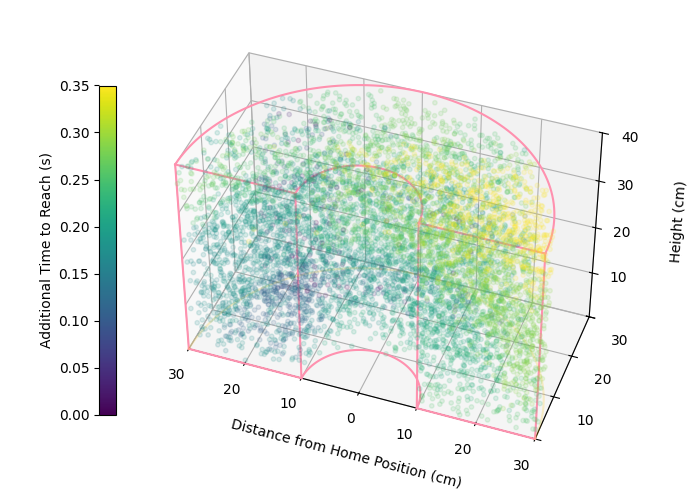}
       \caption{PID 33 XGBoost map of difficulties.}
     \end{subfigure}
     \begin{subfigure} {.3\linewidth}
      \includegraphics[width=\linewidth]{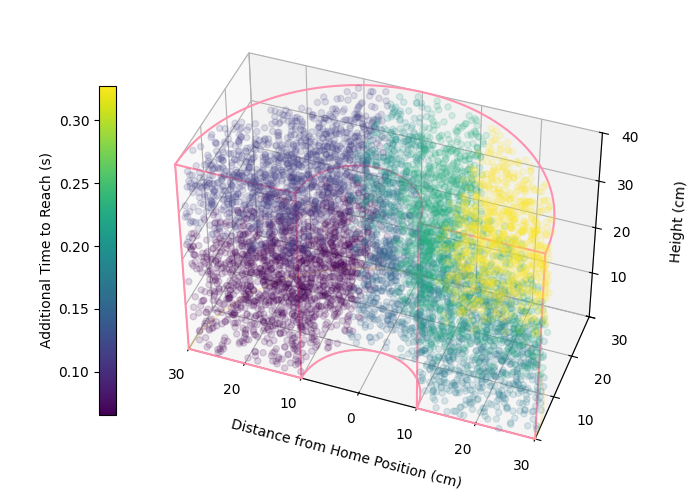}
       \caption{PID 33 causal map of difficulties}
     \end{subfigure}
     \caption{Visual comparison of methods. We show the user's ground truth difference from the neurotypical population (a). Independently estimating the stroke survivor and neurotypical data leads to homogenous difficulty estimates over the workspace (b). Using causal trees instead provides distinctive regions of differences in difficulty that are robust to outliers (c). }
     \label{fig:personalized_difficulty}
\end{figure*}

\subsection{Training Details}
We are interested in estimating the personal functional difficulty of a participant. To evaluate this, we aggregated each user's data across visits, so each participant corresponded to a dataset of up to 300 reaches. We split each dataset into 80\% training and 20\% testing. We used the same neurotypical dataset as the baseline for all participants, reflective of a real-world use case where baseline data are collected prior to system deployment.

\begin{figure}[ht!]
\vspace{-2em}
    \centering
    \includegraphics[width=\linewidth]{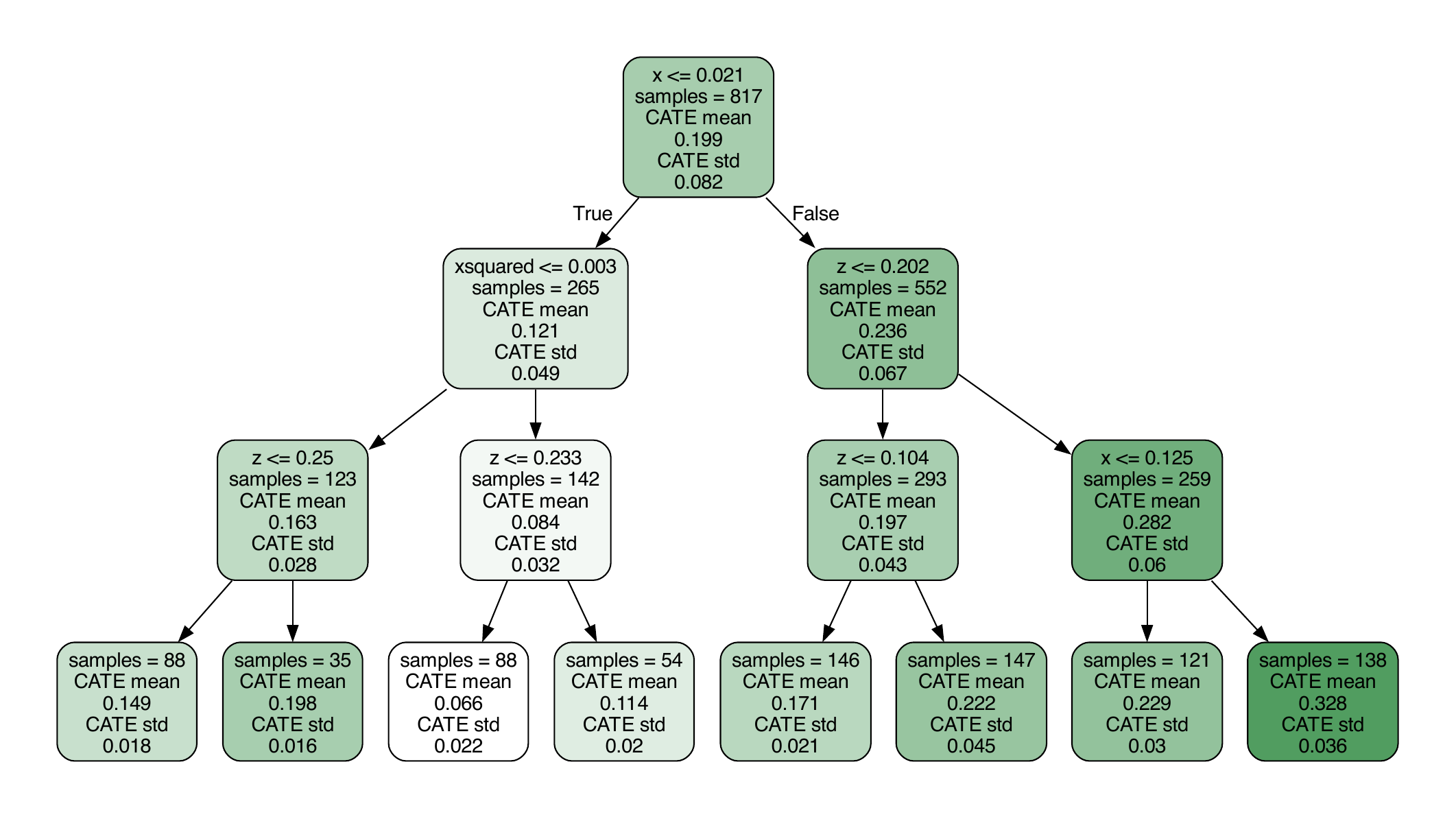}
    \caption{Tree interpretation of the difficulty map for PID 33.}
    \label{fig:tree}
    \vspace{-1.1em}
\end{figure}

We randomly selected five participants to tune hyperparameters for all of the algorithms, and used the other participants to evaluate the algorithms. This ensures we do not leak any information from the training set into the testing set. We present the hyperparameters we used for each of the algorithms we tested in \autoref{table:hyperparams}. We evaluated across 20 random seeds.

\subsection{Personalized Functional Difficulty Estimation}

To evaluate each model, we perform a \textit{within-subject evaluation} and a \textit{aggregated evaluation}. 
We present our results in \autoref{table:results}. 
The \textit{within-subject evaluation} used the mean-squared error (MSE) metric to assess the average difference between predicted and actual functional difficulty, expressed in seconds squared for each participant, we then averaged each participant's MSE to report the final value. 
We found that causal trees have the lowest MSE of all baseline methods, achieving an MSE of .3 $s^2$. 
Our \textit{aggregated evaluation} calculated the $r^2$ value across all participants' predicted and actual functional difficulty, expressed as a proportion of explained variance to total variance. 
Causal trees also perform the better than all the baselines we tested, with an $r^2$ that indicates that 82.6\% of the variance in times are explained with this approach.
These results indicate that causal trees result in better user models compared to non-causal approaches because they can better reason about uncertainty.
We qualitatively compare the causal tree approach and the best performing baseline for PID 33 in \autoref{fig:personalized_difficulty}, showing that causal trees lead to clearer regions of varied difficulties.

\begin{table}[ht]
\caption{Personalized functional difficulty results. Arrows indicate the direction of better performance for each metric.}
\label{table:results}
\adjustbox{width=\linewidth}{
\begin{tabular}{ rcc } 
\hline
Model &  MSE $\downarrow$ ($\pm SE$) & agg. $r^2$ $\uparrow$ ($\pm SE$) \\
\hline
Causal Forest (Ours) & \textbf{.300 $\pm$ .019} & \textbf{ .826 $\pm$ .006} \\
XGBoost & .324 $\pm$ .021 & .812 $\pm$ .006  \\
Random Forest & .337 $\pm$ .021 & .805 $\pm$ .006  \\
Neural Network & .345 $\pm$ .022 & .800 $\pm$ .006  \\
Support Vector Machine & .488 $\pm$ .020 & .718 $\pm$ .008  \\
k-Nearest Neighbors & .380 $\pm$ .024 & .780 $\pm$ .006  \\
Decision Tree & .745 $\pm$ .045 & .571 $\pm$ .011  \\
\hline
\end{tabular}
}
\end{table}

\subsection{Visualization of Difficulty}
\autoref{fig:personalized_difficulty} and \autoref{fig:tree} illustrates two techniques to visualize our approach for estimating individual difficulty metrics.
In the workspace view (\autoref{fig:personalized_difficulty}), we show the regions of similar exercise difficulty spatially. These areas are distinctive, which allows clinicians or rehabilitation robots to sample from a particular region in the space to manage difficulty while providing slightly different exercises.

The tree view (\autoref{fig:tree}) is an explainable diagram that shows how the causal tree splits regions of the workspace based on the user's performance. Each node is colored with the difficulty level; darker colors indicate more difficult exercise regions. Each level of the tree shows different granularities of exercise difficulty. For example, if a rehabilitation specialist only cares about two levels of difficulty, they can examine the second level of the tree, but if they want to reason about eight levels of difficulty, they can examine the fourth level of the tree.

\section{Discussion}

This work presented an algorithm that estimates nominal and functional exercise difficulty given a set of exercise parameters and outcome measures. We evaluated this technique on a reaching task performed by both neurotypical participants and post-stroke participants. We found that our method results in more accurate estimations of personalized functional difficulty.

A key benefit of this approach is that it can be applied to other kinds of exercises and outcome measures to estimate difficulty and facilitate adaptation. For example, previous work that adapts tasks based on physiological features \cite{shirzad2013adaptation} or game performance \cite{gorvsivc2017comparison} could leverage this approach to select specific exercises with target difficulty levels. This additionally provides a principled technique for determining difficulty for a variety of tasks, which can encourage the development of new exercises that have various adaptable characteristics, such as in serious games \cite{gorvsivc2017comparison}.

\textbf{Limitations.} We evaluated our algorithm on a difficulty value based on time to reach. However, there are many metrics that are important for assessing difficulty. For example, movement quality is typically a metric of interest for clinicians and users with limited mobility \cite{winstein2019dosage}. Future work may leverage computer vision techniques to predict movement quality as a separate metric and frame personalized functional difficulty modeling as a multi-outcome causal inference problem. Other future work may design metrics that factor both time and quality into a single value to use as an outcome metric.

We additionally only evaluated our difficulty estimation technique on a reaching task. Other tasks may involve other numerical exercise characteristics, such as stiffness for an arm that users push against, which may be more difficult to visualize for users. Future work may evaluate personalized functional difficulty modeling on different tasks, as well as iterate over data visualizations to understand how to effectively communicate these difficulty models to users and clinicians. 

\textbf{Conclusion.} We presented a technique for learning individualized models of user difficulty based on causal trees. We found that jointly modeling neurotypical and stroke survivor performance better estimates exercise difficulty compared to separately estimating these quantities, which we evaluated across several stroke survivors. Causal trees provide readily interpretable visualizations that can aid in communication between clinicians and stroke survivors. We hope that this technique opens new directions for adaptable and personalized rehabilitation practices. 

\appendix

We provide the code used in this experiment at \href{https://github.com/interaction-lab/stroke-exercise-difficulty-estimation}{\texttt{https://github.com/interaction-lab/stroke-\\exercise-difficulty-estimation}}.

\bibliographystyle{IEEEtran}
\bibliography{main}

\end{document}